\documentclass[times,twocolumn,final]{elsarticle}

\usepackage{medima}
\usepackage{framed,multirow}
\usepackage[utf8]{inputenc} 
\usepackage[T1]{fontenc}    
\usepackage{amsfonts}
\usepackage{nicefrac}
\usepackage[final,nopatch=footnote]{microtype}
\usepackage{stfloats}
\usepackage{csquotes}
\usepackage{xspace}

\usepackage{amssymb}
\usepackage{latexsym}
\usepackage{textcomp}
\usepackage{gensymb}

\usepackage{url}
\usepackage{xcolor}

\usepackage[%
breaklinks=true,
colorlinks=true,
bookmarksnumbered=true
]{hyperref}
\usepackage[labelfont=bf,font=small]{caption}

\usepackage{booktabs}

\usepackage[official]{eurosym}
\usepackage{pdfpages}
\usepackage{stackengine}
\usepackage{graphicx}
\usepackage{lscape}

\usepackage{booktabs}
\usepackage{float}
\usepackage{framed,multirow}
\usepackage{setspace}
\usepackage{threeparttable}
\usepackage{tabularx}
\usepackage{amsmath}
\usepackage{siunitx}
\DeclareSIUnit\px{px}

\newcolumntype{L}[1]{>{\raggedright\arraybackslash}p{#1}} 
\newcolumntype{C}[1]{>{\centering\arraybackslash}p{#1}} 
\newcolumntype{R}[1]{>{\raggedleft\arraybackslash}p{#1}} 

\usepackage{amssymb}
\usepackage{latexsym}

\usepackage{booktabs}
\usepackage{enumitem}
\usepackage{ifthen}

\usepackage[official]{eurosym}
\usepackage{pdfpages}
\usepackage{stackengine}
\usepackage{algorithm}
\usepackage{algorithmic}
\usepackage[labelformat=simple]{subcaption}
\makeatletter
\renewcommand\p@subfigure{\thefigure\,}
\makeatother

\usepackage{mathrsfs}

\usepackage{acro}
\acsetup{make-links=true}
\newcommand{\acrocolor}[1]{\textcolor{black}{#1}}
\DeclareAcronym{hsi}{short=\acrocolor{HSI},long=hyperspectral imaging}
\DeclareAcronym{msi}{short=\acrocolor{MSI},long=multispectral imaging}
\DeclareAcronym{tpi}{short=\acrocolor{TPI},long=tissue parameter images}
\DeclareAcronym{dsc}{short=\acrocolor{DSC},long=dice similarity coefficient}
\DeclareAcronym{asd}{short=\acrocolor{ASD},long=average surface distance}
\DeclareAcronym{nsd}{short=\acrocolor{NSD},long=normalized surface dice}
\DeclareAcronym{sto2}{short=\acrocolor{\ce{StO2}},long=tissue oxygen saturation}
\DeclareAcronym{npi}{short=\acrocolor{NPI},long=near-infrared perfusion index}
\DeclareAcronym{twi}{short=\acrocolor{TWI},long=tissue water index}
\DeclareAcronym{thi}{short=\acrocolor{THI},long=tissue hemoglobin index}
\DeclareAcronym{elu}{short=\acrocolor{ELU},long=exponential linear unit}
\DeclareAcronym{gpu}{short=\acrocolor{GPU},long=graphics processing unit}
\DeclareAcronym{sd}{short=\acrocolor{SD},long=standard deviation}
\DeclareAcronym{ce}{short=\acrocolor{CE},long=cross-entropy}
\DeclareAcronym{json}{short=\acrocolor{JSON}, long=JavaScript object notation}
\DeclareAcronym{tsne}{short=\acrocolor{$t$-SNE}, long=$t$-distributed stochastic neighbour approach}
\DeclareAcronym{svm}{short=\acrocolor{SVM}, long=support vector machine}
\DeclareAcronym{ood}{short=\acrocolor{OOD},long=out-of-distribution}
\DeclareAcronym{iqr}{short=\acrocolor{IQR},long=interquartile range}
\DeclareAcronym{soa}{short=\acrocolor{SOA},long=state-of-the-art}

\definecolor{mialinkcolor}{RGB}{0,128,172}
\newcommand{\autocite}[1]{\citep{#1}}

\newcommand{\dataset}[1]{\textit{#1}}

\newcommand{\varTotalImages}{600\xspace}
\newcommand{\varTotalPigs}{33\xspace}

\newcommand{\varTotalPigsOld}{20\xspace}
\newcommand{\varTotalImagesIsolationReal}{94\xspace}
\newcommand{\varTotalPigsIsolationReal}{25\xspace}
\newcommand{\varTotalImagesGlove}{142\xspace}
\newcommand{\varTotalImagesNoGlove}{364\xspace}

\newcommand{\varTotalTrainingPigs}{15\xspace}
\newcommand{\varTotalTrainingImages}{340\xspace}
\newcommand{\varTotalTestPigsInDistribution}{5\xspace}
\newcommand{\varTotalTestImagesInDistribution}{166\xspace}
\newcommand{\varTotalClasses}{19\xspace}
\newcommand{\varTotalOrganClasses}{18\xspace}

\newcommand{\varDSCBaselineHSIInDistribution}{0.86 (\ac{sd} 0.10)\xspace}

\newcommand{\varDSCBaselineRGBInDistribution}{0.83 (\ac{sd} 0.10)\xspace}

\newcommand{\varHSIDropRange}{\SIrange{5}{45}{\percent}\xspace}
\newcommand{\varRGBDropRange}{\SIrange{10}{46}{\percent}\xspace}
\newcommand{\varHSIImprovementRangeDSC}{\SIrange{9}{90}{\percent}\xspace}
\newcommand{\varRGBImprovementRangeDSC}{\SIrange{9}{67}{\percent}\xspace}
\newcommand{\varHSIImprovementRangeNSD}{\SIrange{16}{96}{\percent}\xspace}
\newcommand{\varRGBImprovementRangeNSD}{\SIrange{15}{79}{\percent}\xspace}
\newcommand{\varHSIImprovementMax}{\SI{90}{\percent}\xspace}
\newcommand{\varRGBImprovementMax}{\SI{67}{\percent}\xspace}
\newcommand{\varHSIMaxDrop}{\SI{45}{\percent}\xspace}
\newcommand{\varRGBMaxDrop}{\SI{46}{\percent}\xspace}

\newcommand{\varGallbladderLiverNeigbor}{\SI{83.9}{\percent}\xspace}
\newcommand{\varMajorVeinPeritoneum}{\SI{60.1}{\percent}\xspace}
\newcommand{\varHSIDropAverage}{\SI{23}{\percent}\xspace}
\newcommand{\varRGBDropAverage}{\SI{30}{\percent}\xspace}

\journal{Medical Image Analysis}
\verso{Seidlitz \& Sellner \textit{et~al.}}

\begin{document}
\hypersetup{allcolors=mialinkcolor}

\begin{frontmatter}

\title{Handling Geometric Domain Shifts in Semantic Segmentation of Surgical RGB and Hyperspectral Images}%

\author[1,2,3,4]{Silvia \snm{Seidlitz}}\corref{cor1}
\ead{s.seidlitz@dkfz-heidelberg.de}
\author[1,2,3,4]{Jan \snm{Sellner}\corref{cor1}}
\ead{j.sellner@dkfz-heidelberg.de}
\author[4,5,6]{Alexander \snm{Studier-Fischer}}
\author[1]{Alessandro \snm{Motta}}
\author[5,6]{Berkin \snm{\"Ozdemir}}
\author[5,6]{Beat P. \snm{M\"uller-Stich}}
\author[2,5,6]{Felix \snm{Nickel}}
\author[1,2,3,4,6]{Lena \snm{Maier-Hein}}
\cortext[cor1]{First authors (*) contributed equally to this paper and the order was assigned randomly.}

\address[1]{Division of Intelligent Medical Systems (IMSY), German Cancer Research Center (DKFZ), Heidelberg, Germany}
\address[2]{Helmholtz Information and Data Science School for Health, Karlsruhe/Heidelberg, Germany}
\address[3]{Faculty of Mathematics and Computer Science, Heidelberg University, Heidelberg, Germany}
\address[4]{National Center for Tumor Diseases (NCT), NCT Heidelberg, a partnership between DKFZ and university medical center Heidelberg}
\address[5]{Department of General, Visceral, and Transplantation Surgery, Heidelberg University Hospital, Heidelberg, Germany}
\address[6]{Medical Faculty, Heidelberg University, Heidelberg, Germany}

\received{2 August 2024}
\finalform{2 August 2024}
\accepted{2 August 2024}
\availableonline{2 August 2024}


\begin{abstract}
Robust semantic segmentation of intraoperative image data holds promise for enabling automatic surgical scene understanding and autonomous robotic surgery. While model development and validation are primarily conducted on idealistic scenes, geometric domain shifts, such as occlusions of the situs, are common in real-world open surgeries. To close this gap, we (1) present the first analysis of \ac{soa} semantic segmentation models when faced with geometric \ac{ood} data, and (2) propose an augmentation technique called \enquote{Organ Transplantation}, to enhance generalizability. Our comprehensive validation on six different \ac{ood} datasets, comprising \varTotalImages RGB and \ac{hsi} cubes from \varTotalPigs pigs, each annotated with \varTotalClasses classes, reveals a large performance drop in \ac{soa} organ segmentation models on geometric \ac{ood} data. This performance decline is observed not only in conventional RGB data (with a \ac{dsc} drop of \varRGBMaxDrop) but also in \ac{hsi} data (with a \ac{dsc} drop of \varHSIMaxDrop), despite the richer spectral information content. The performance decline increases with the spatial granularity of the input data. Our augmentation technique improves \ac{soa} model performance by up to \varRGBImprovementMax for RGB data and \varHSIImprovementMax for \ac{hsi} data, achieving performance at the level of in-distribution performance on real \ac{ood} test data. Given the simplicity and effectiveness of our augmentation method, it is a valuable tool for addressing geometric domain shifts in surgical scene  segmentation, regardless of the underlying model. Our code and pre-trained models are publicly available at \href{https://github.com/IMSY-DKFZ/htc}{https://github.com/IMSY-DKFZ/htc}.
\end{abstract}

\end{frontmatter}

\acresetall

\begin{figure*}[htb]
    \centering
    \includegraphics[width=\textwidth]{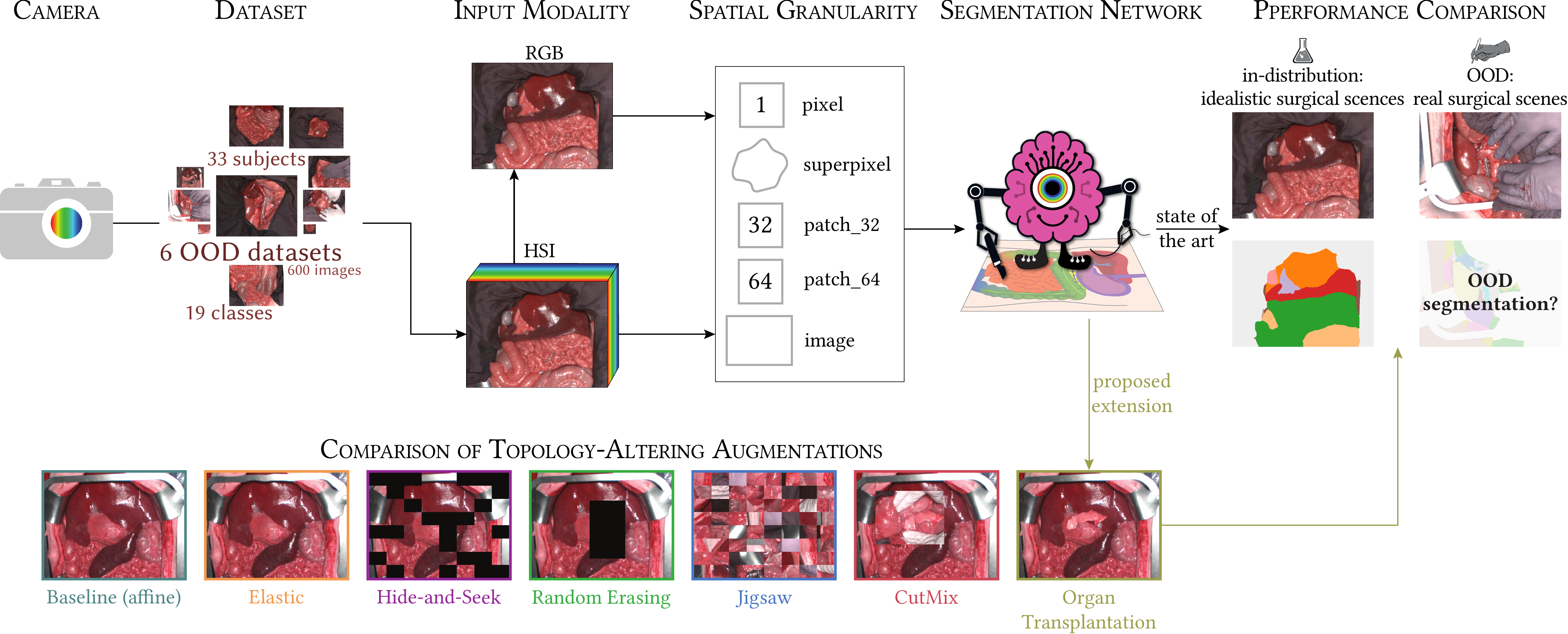}
    \caption{\textbf{Handling geometric domain shifts in the semantic segmentation of open surgery images.} We address the generalizability of surgical scene segmentation algorithms when faced with \acf*{ood} geometries for two modalities (RGB and \acf*{hsi}) and different spatial granularities of the input data (from pixels to images). Our proposed Organ Transplantation augmentation method outperforms topology-altering augmentation techniques adapted from the general computer vision community and yields performance on par with in-distribution performance.}
    \label{fig:intro}
\end{figure*}

\section{Introduction}

To achieve context-aware assistance and autonomous robotic surgery, automated surgical scene segmentation is crucial. Recent studies have demonstrated that deep learning-based surgical scene segmentation can achieve high accuracy \autocite{kadkhodamohammadi_feature_2019, scheikl_deep_2020} up to the level of human performance when using \ac{hsi} instead of RGB data. Additionally, \ac{hsi} provides the advantage of offering functional tissue information \autocite{seidlitz_robust_2022}. However, the important issue of geometric domain shifts, commonly encountered in real-world surgical scenes due to variations in procedures or situs occlusions (cf. \autoref{fig:intro}), has not been addressed in the literature. It remains uncertain whether the \ac{soa} image-based segmentation networks presented in \autocite{seidlitz_robust_2022} can generalize to \ac{ood} geometries. It has recently been shown that algorithms for surgical instrument segmentation fail to generalize to unseen surgery types involving known instruments in unfamiliar contexts \autocite{kitaguchi_limited_2022}. Furthermore, image-based segmentation models demonstrated inferior generalization towards unseen surgeries than pixel-based segmentation \autocite{seidlitz_robust_2022}. However, to our knowledge, there has been no investigation or methodological contribution addressing geometric domain shifts in the context of surgical scene segmentation.

Domain shifts are being intensively studied in the general deep learning community, with data augmentation emerging as a simple yet powerful technique to improve generalizability. Traditional augmentation approaches can be categorized into geometric (e.g., rotating), photometric (e.g., color jittering), noise (e.g., Random Erasing), kernel (e.g., Gaussian blur) and image mixing (e.g., CutMix) transformations \autocite{shorten_survey_2019, alomar_data_2023}. Geometric transformations are the most common augmentation method in deep learning-based semantic image segmentation \autocite{kar_review_2021}. Surgical applications are no exception. Our analysis of the state of the art (35 publications on tissue or instrument segmentation) predominantly identified the use of geometric, photometric, and kernel transformations. Only a single publication \autocite{ros_comparative_2021} mentioned the use of elastic transformations and Random Erasing, where a rectangular area of the image is blacked out \autocite{zhong_random_2020}. Likewise, augmentation schemes for tissue classification using \ac{hsi} data have so far been limited to geometric augmentations. Topology-altering transformations such as Hide-and-Seek (divides the image into a grid of patches that are randomly blacked out) \autocite{singh_hide-and-seek_2017}, Jigsaw (divides images into grids of patches that are randomly exchanged across images) \autocite{chen_image_2019}, CutMix (copies a random patch from one image to another) \autocite{yun_cutmix_2019} and CutPas (copies an object onto a random background image) \autocite{dwibedi_cut_2017} (cf. \autoref{fig:intro} for illustrations) have been proposed for image classification and object detection. However, to the best of our knowledge, their potential for surgical scene segmentation has not yet been explored.

To address these gaps in the literature, this paper investigates the following research questions:
\begin{enumerate}[noitemsep, leftmargin=1cm]
    \item[RQ1:] What is the impact of geometric domain shifts on the performance of state-of-the-art RGB and \ac{hsi} surgical scene segmentation models?
    \item[RQ2:] How does the spatial granularity of the input data affect the degree of performance degradation?
    \item[RQ3:] Can topology-altering augmentation techniques compensate for geometric domain shifts?
\end{enumerate}

Although portions of this work were previously published in \autocite{sellner_context_2023}, this paper introduces several new contributions:
\begin{enumerate}[noitemsep, leftmargin=1cm]
    \item \textit{Additional experiments:} We conducted new experiments to investigate performance degradation in the presence of geometric domain shifts as a function of the input spatial granularity (pixels, superpixels, patches and images).
    \item \textit{In-depth results:} We added an analysis of common organ neighborhood relations from a set of \varTotalTestImagesInDistribution semantically annotated images. This new analysis contributes to the understanding of which organ classes are most affected by geometric domain shifts. Furthermore, we included exemplary segmentations of our models.
    \item \textit{Comprehensive description of our work:} We substantially expanded our Materials and Methods (\autoref{sec:methods}), Experiments (\autoref{sec:experiments}), Results (\autoref{sec:results}), and Discussion (\autoref{sec:discussion}) sections by including additional information, thereby improving the comprehensibility and reproducibility of our work.
\end{enumerate}

\section{Materials and Methods}
\label{sec:methods}

The following sections describe the hardware and data used in this work (Section 2.1) and the deep learning-based semantic scene segmentation pipeline (Section 2.2). Code and pretrained models are available in our GitHub repository\footnote{\href{https://github.com/IMSY-DKFZ/htc}{https://github.com/IMSY-DKFZ/htc}} \autocite{sellner_htc_2023}.

\subsection{Image Acquisition and Annotation}
\label{sec:data}

The data was collected at Heidelberg University Hospital following approval by the Committee on Animal Experimentation of the regional council of Baden-W\"urttemberg in Karlsruhe, Germany (G-161/18 and G-262/19). The medical device-graded \ac{hsi} camera system Tivita\textsuperscript{\textregistered} Tissue (Diaspective Vision GmbH, Am Salzhaff, Germany) was used, capturing 100 spectral channels in the range between \SI{500}{\nm} and \SI{1000}{\nm} at a spectral resolution of \SI{5}{\nm} and spatial resolution of $640 \times 480$ pixels. RGB images were reconstructed from the \ac{hsi} cubes by aggregating spectral channels in the blue, green, and red ranges \autocite{holmer_hyperspectral_2018}. The camera system captures an area of approximately $20 \times \SI{30}{\cm}$. An integrated distance calibration system, comprising two light marks that overlap when the correct measurement distance of about \SI{50}{\cm} is maintained, ensured consistent imaging distances. To prevent spectral distortion from stray light, all light sources except the integrated halogen lighting unit were turned off during image acquisition, and the window blinds were closed. The acquisition of one \ac{hsi} cube took approximately seven seconds \autocite{holmer_hyperspectral_2018, kulcke_compact_2018}.

A total of \varTotalImages intraoperative \ac{hsi} cubes from \varTotalPigs pigs were collected and semantically annotated with \varTotalClasses classes. Two annotators performed the semantic annotations using vector annotation tools on the SuperAnnotate platform (SuperAnnotate, Sunnyvale, USA)\footnote{\href{https://superannotate.com/}{https://superannotate.com/}}. To ensure correctness and consistency, a third medical expert subsequently reviewed all annotations. The annotations include two thoracic organs (heart, lung), eight abdominal organs (stomach, small bowel, colon, liver, gallbladder, pancreas, kidney, spleen), and one pelvic organ (bladder). For the kidney, images were taken before and after the removal of Gerota’s fascia, and labeled \enquote{kidney with Gerota’s fascia} and \enquote{kidney}, respectively. Additionally, subcutaneous fat, skin, muscle tissue, omentum, peritoneum, and major veins were annotated. Pixels corresponding to any inorganic object (e.g., cloth, metallic objects, and gloves) were labeled as background.

\subsection{Deep Learning-Based Semantic Scene Segmentation}
\label{sec:dl}

Our contribution is based on the hypothesis that application-specific data augmentation can effectively address geometric domain shifts. Instead of altering the network architecture of previously successful segmentation methods, we thus focused on combining \ac{soa} segmentation models with a topology-altering data augmentation.

\begin{figure*}[ht]
    \centering
    \includegraphics[width=0.75\textwidth]{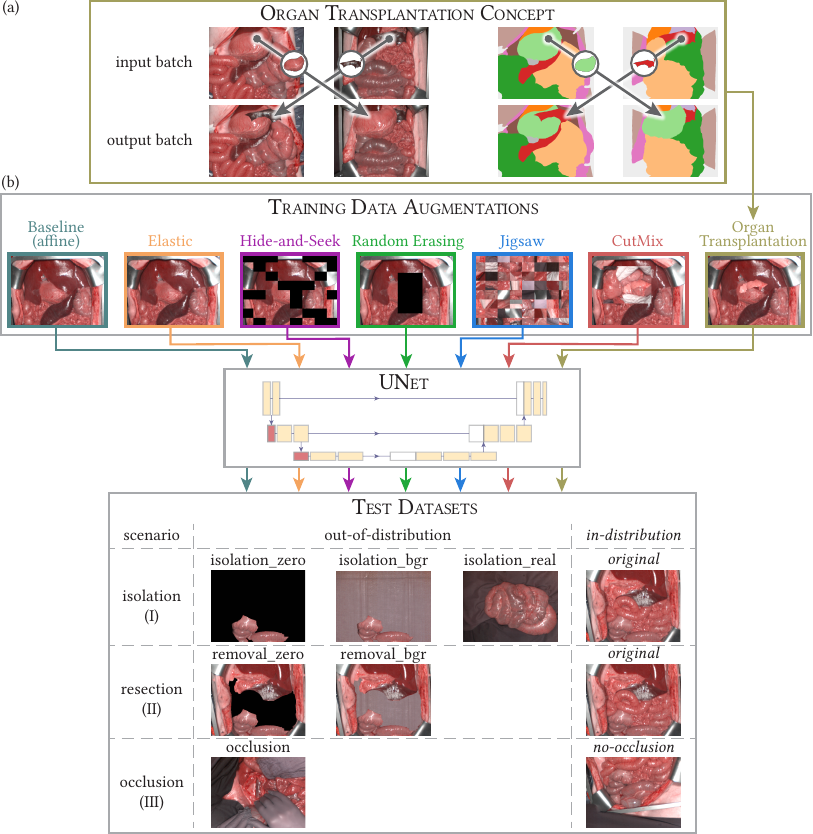}
    \caption{\textbf{Concept of our Organ Transplantation augmentation and experimental setup.} (a) Our Organ Transplantation augmentation involves transferring image features and corresponding segmentations of randomly selected organs (here: stomach and spleen) between images within a batch. (b) We compare the generalization performance of state-of-the-art organ segmentation models under geometric domain shifts upon equipping the models with either the proposed new approach or one of the six adapted data augmentation techniques. Our test datasets encompass the geometric out-of-distribution scenarios (I) organs in isolation, (II) organ resections, and (III) situs occlusions, in addition to in-distribution data. This figure is adapted from \autocite{sellner_context_2023}.}
    \label{fig:concept}
\end{figure*}

\textit{Surgery-inspired data augmentation:} Similar to how a donor organ is received during transplantation, our Organ Transplantation augmentation technique copies all pixels of a specific object class (e.g., an organ or background) onto a different surgical scene. As shown in \autoref{fig:concept}, the corresponding object segmentation is copied and pasted accordingly. This approach places the organ in an unusual geometric context while preserving its shape and texture. The augmentation was inspired by the CutPas image-mixing augmentation initially proposed for object detection \autocite{dwibedi_cut_2017}. It has since been adapted for instance segmentation \autocite{ghiasi_simple_2021} and for generating low-cost datasets through image synthesis from a few real-world images in surgical instrument segmentation \autocite{wang_rethinking_2022}.

\textit{Further data augmentation methods:} Several other topology-altering augmentation methods (cf. \autoref{fig:concept}) could potentially improve generalizability under geometric domain shifts. Hide-and-Seek and Random Erasing are noise augmentations that obscure all pixels within rectangular regions of an image, thereby creating artificial situs occlusions. In contrast, the image-mixing techniques Jigsaw and CutMix transfer all pixels from rectangular regions of one image onto a different surgical scene. For our segmentation task, we modified these augmentations by also copying and pasting the corresponding segmentations. As a result, the adapted Jigsaw and CutMix augmentations not only occlude parts of the original scene but also place image segments in unusual contexts. \\

The \ac{soa} models and pipeline are detailed in \autocite{seidlitz_robust_2022}. Only a brief summary of model architectures, data pre-processing and training setup is given here.

\textit{\ac{soa} model architectures:} While the same model architectures were utilized for both \ac{hsi} and RGB-based segmentation, different architectures were proposed for various spatial granularities. For pixel-based segmentation, each spectrum was processed individually using a one-dimensional convolutional classification network. Superpixels, which were precomputed from the RGB image and resized to a fixed shape, were input into an EfficientNet-B5 encoder \autocite{tan_efficientnet_2019} that assigned a class label to each superpixel. For patch-based segmentation, two models were suggested: the patch\_32 model, which uses $32 \times 32 \times c$ (width $\times$ height $\times$ number of spectral channels $c$) patches, and the patch\_64 model, which uses $64 \times 64 \times c$ patches. Both the patch\_32 and patch\_64 models use a U-Net \autocite{ronneberger_u-net_2015} with an EfficientNet-B5 encoder for segmentation. For all pixels, superpixels and patches composing an image, predictions were collected to obtain the image segmentation. For image-based segmentation, the entire image was processed by a model using a U-Net with an EfficientNet-B5 encoder, which directly provided the segmentation output. In all instances where the EfficientNet-B5 encoder was used, weights were pre-trained on the ImageNet data.

\textit{\ac{hsi} pre-processing:} To eliminate sensor noise and convert the acquired \ac{hsi} data from radiance to reflectance, the raw \ac{hsi} cubes were automatically corrected using a pre-recorded white and dark reference cube, as described in Holmer et al. (2018). Each pixel spectrum was $\ell^1$-normalized to account for multiplicative illumination changes. 

\textit{Training setup:} To avoid biases from differences in the training setup and enable a fair comparison of modalities, spatial granularities and augmentation methods, the training setup was made as comparable as possible across all models. The same model hyperparameter settings as described in the \ac{soa} were used \autocite{seidlitz_robust_2022}: Before extracting pixels, superpixels, or patches, images were augmented using the geometric transformations shift (limit of 0.0625), scale (limit of 0.1), and rotate (limit of \SI{45}{\degree}), each applied with a probability of 0.5. The Adam optimization algorithm \autocite{kingma_adam_2017} was used in conjunction with an exponential learning rate schedule (initial learning rate: 0.001, decay rate $\gamma$: 0.99, Adam decay rates $\beta_1$: 0.9 and $\beta_2$: 0.999). The Dice loss and cross-entropy loss were weighted equally to calculate the loss function. Training was conducted for 100 epochs with an epoch size of 500 images, and stochastic weight averaging \autocite{izmailov_averaging_2018} was applied during the final 10 epochs. Dropout regularization was applied with a probability of 0.1.

\section{Experiments}
\label{sec:experiments}

To examine the performance of \ac{soa} surgical scene segmentation models under geometric domain shifts as a function of input modality (RQ1) and spatial granularity (RQ2), and assess the generalizability improvements provided by topology-altering augmentations (RQ3), we addressed the following \ac{ood} scenarios:

\begin{enumerate}
    \item \textit{Organs in isolation:} Abdominal linens are routinely used to control excessive bleeding, absorb blood and secretions and protect soft tissue and organs. Certain surgeries (e.g., enteroenterostomy) necessitate covering all organs except one, creating the need to robustly identify an organ without any information about neighboring organs.
    \item \textit{Organ resections:} In surgical resections, organ parts or even the entire organ are removed, necessitating the identification of surrounding organs despite the absence of a common neighboring organ.
    \item \textit{Situs occlusions:} Large portions of the surgical site can be obscured by the procedure itself, introducing \ac{ood} neighbors such as gloved hands, which challenge the correct segmentation of the visible parts of the site.
\end{enumerate}

\textit{Real-world datasets:}
We acquired a total of \varTotalImages intraoperative \ac{hsi} cubes from \varTotalPigs pigs. Out of these, \varTotalImagesIsolationReal images from \varTotalPigsIsolationReal pigs compose the dataset \dataset{isolation\_real}. In these images, all organs except a specific one were covered by abdominal linen, providing isolation images for all \varTotalOrganClasses organ classes. To examine the impact of occlusions, we collected \varTotalImagesGlove images of \varTotalPigsOld pigs with real-world situs occlusions by gloved hands (dataset \dataset{occlusion}) and \varTotalImagesNoGlove images without such occlusions (dataset \dataset{no-occlusion}). The dataset \dataset{original} is the combination of the datasets \dataset{occlusion} and \dataset{no-occlusion}. Example images for all datasets are illustrated in \autoref{fig:concept}.

\textit{Manipulated datasets:}
We enhanced our real-world datasets with four manipulated datasets. For a simulated organs in isolation scenario, we iterated over all images $I$ and labels $l$ in the dataset \dataset{original} and replaced all pixels in $I$ except those corresponding to class $l$ with either zeros to create the dataset \dataset{isolation\_zero}, or with spectra from a background image to form the dataset \dataset{isolation\_bgr}. Similarly, we generated the resection datasets \dataset{removal\_zero} and \dataset{removal\_bgr} by replacing all pixels in $I$ corresponding to class $l$ with zeros and background spectra, respectively. Example images for our manipulated datasets are illustrated in \autoref{fig:concept}.

\textit{Train-test-split:}
The dataset \dataset{original} was previously used for developing the \ac{soa} models in \autocite{seidlitz_robust_2022}. The same split as described in the publication was employed, which includes a hold-out test split of \varTotalTestImagesInDistribution images from \varTotalTestPigsInDistribution pigs and a training split of \varTotalTrainingImages images from \varTotalTrainingPigs pigs. To ensure a fair comparison across models and \ac{ood} scenarios, the same train-test-split at the pig level was consistently applied: The test splits of \dataset{isolation\_zero}, \dataset{isolation\_bgr}, \dataset{removal\_zero}, and \dataset{removal\_bgr} were created by manipulating the images in the test split of \dataset{original}. In the occlusion scenario, models were trained on the subset of images in the train split of \dataset{original} that do not contain occlusions, and testing was performed on the subset of the test split of \dataset{original} without occlusions (\dataset{no-occlusion}) and with occlusions (\dataset{occlusion}). As described in \autoref{sec:dl}, most model hyperparameters were set according to the \ac{soa}. Only hyperparameters related to the topology-altering augmentations, specifically the probability $p$ of applying the augmentation, were optimized using a grid search with $p \in \{ 0.2, 0.4, 0.6, 0.8, 1 \}$. The optimal $p$-value was determined using five-fold-cross-validation on the train splits of the datasets \dataset{original}, \dataset{isolation\_zero}, and \dataset{isolation\_bgr}.

\textit{Validation strategy:}
To overcome limitations of individual metrics \autocite{reinke2024understanding}, we assessed the performance for each class label $l$ using both the overlap-based metric \ac{dsc} \autocite{dice_measures_1945} and the boundary-based metric \ac{nsd} \autocite{nikolov_deep_2021}, as recommended by the Metrics Reloaded framework \autocite{maier-hein_metrics_2023}. For the \ac{nsd}, we used the class-specific thresholds reported in the \ac{soa} \autocite{seidlitz_robust_2022}. Metric aggregation adhered to the hierarchical structure of the data by first macro-averaging the class-level metric value $M_{l}$ ($M \in \{ \operatorname{DSC}, \operatorname{NSD}\}$) across all images of a single pig, and then averaging across pigs. In the organ removal scenario, for each class label $l$ in an image $I$, a set of metric scores $\{ M_{l} (\hat{l}) \}$ was obtained for the one-by-one removal of all other classes $\hat{l}$ in $I$. To evaluate the effect of removing the most important neighbor of $l$, we selected the smallest score in $\{ M_{l} (\hat{l}) \}$ prior to proceeding with the hierarchical aggregation.

\textit{Performance ranking:}
To compare the performance of our Organ Transplantation augmentation with other topology-altering augmentations, we computed performance rankings and assessed ranking stability following the guidelines in \autocite{wiesenfarth_methods_2021}: We calculated the rank for each augmentation method across 1000 bootstrap samples. For each class label $l$ in each bootstrap sample, we randomly selected $N_l$ subject-level scores without replacement, where $N_l$ is the total number of subjects with images available for class $l$. Hierarchical aggregation of the metric values was performed for each bootstrap sample, resulting in a set of 1000 class-averaged scores that capture the variability across subjects.

\section{Results}
\label{sec:results}

A primary purpose of our study was to examine how geometric domain shifts affect the performance of \ac{soa} segmentation models as a function of (1) the modalities \ac{hsi} and RGB, and (2) various input spatial granularities such as pixels, superpixels, patches and images. The results of this analysis are presented in \autoref{sec:performance_drop}. \autoref{sec:augmentation_results} presents the performance analysis of our Organ Transplantation model, and a comparison to topology-altering augmentations adapted from the general computer vision community is given in \autoref{sec:ranking}.

\subsection{Effects of Geometric Domain Shifts on State-Of-The-Art Surgical Scene Segmentation}
\label{sec:performance_drop}

\begin{figure*}[htb]
    \centering
    \includegraphics[width=1\textwidth]{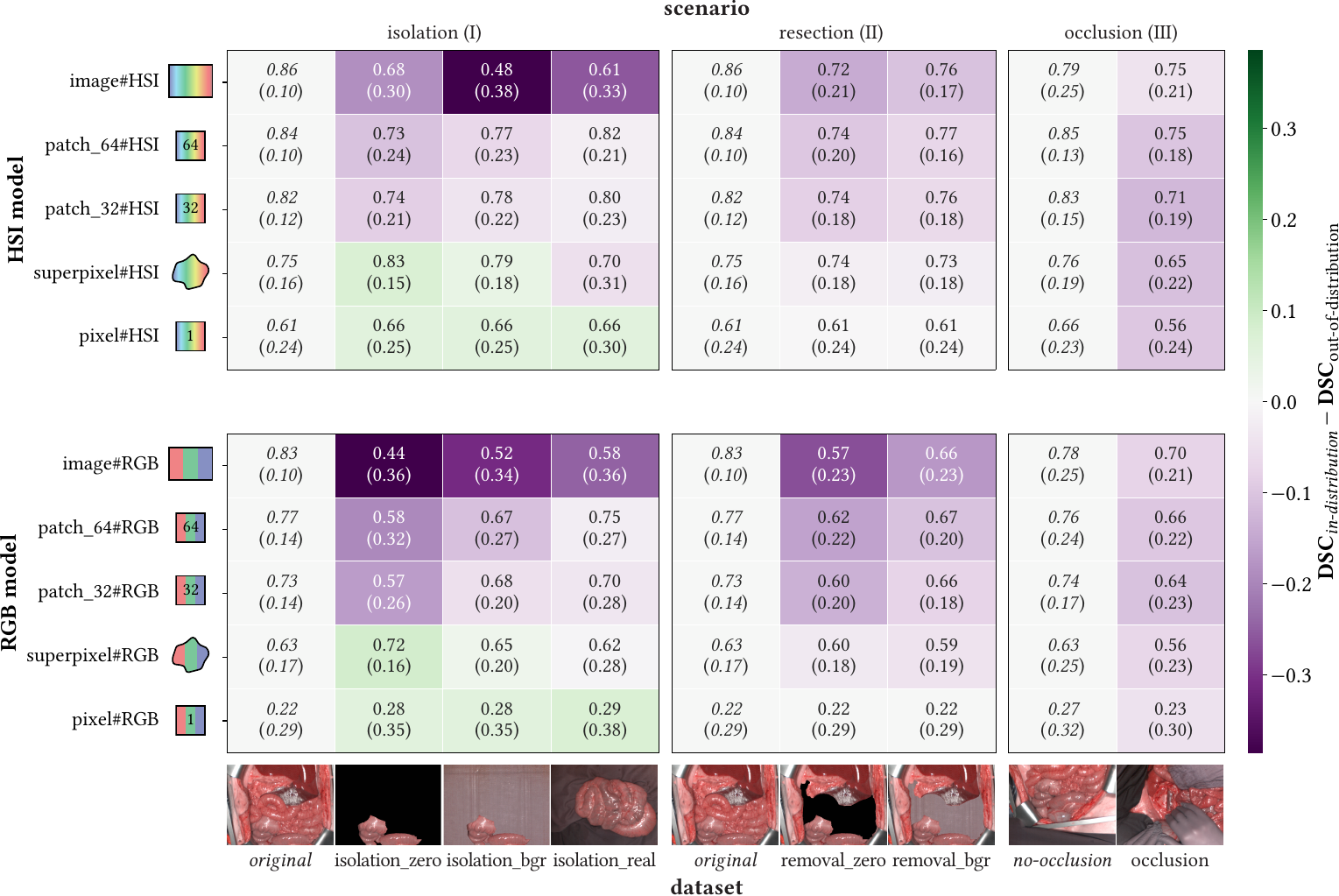}
    \caption{\textbf{Performance degradation in response to geometric domain shifts increases with an increasing spatial granularity.} The segmentation performance on our in-distribution (highlighted in \textit{italic}) and \acf*{ood} scenarios (columns) is shown as a function of the input modality and spatial granularity (rows; top: \acf*{hsi}, bottom: RGB). The numbers denote the average class \acf*{dsc}, with standard deviations across classes indicated in brackets. The color-coding is based on the change in \acs*{dsc} relative to the in-distribution \acs*{dsc} of the same model. Hierarchical aggregation was performed first across all images of a single subject, then across subjects, and finally across class-level \acs*{dsc} scores.}
    \label{fig:spatial_granularities_dsc}
\end{figure*}

\autoref{fig:spatial_granularities_dsc} provides an overview of the segmentation performance, measured by the \ac{dsc}, for all investigated models on the two in-distribution datasets (\dataset{original} and \dataset{no-occlusion}) and six \ac{ood} datasets (\dataset{isolation\_zero}, \dataset{isolation\_bgr}, \dataset{isolation\_real}, \dataset{removal\_zero},\dataset{removal\_bgr}, and \dataset{occlusion}). 

\textit{Performance drop as a function of modality:}
The average performance drop on \ac{ood} data is smaller for \ac{hsi} (\varHSIDropAverage) compared to RGB (\varRGBDropAverage).

\textit{Performance drop as a function of spatial granularity:}
For both modalities, the in-distribution performance is highest for image-based segmentation (RBG: \ac{dsc} of \varDSCBaselineRGBInDistribution; \ac{hsi}: \ac{dsc} of \varDSCBaselineHSIInDistribution) and decreases with reduced input spatial granularity. While pixel-based segmentation models show the lowest overall performance, they do not experience a decline in performance on \ac{ood} data for the isolation and removal scenarios. As the input spatial granularity increases, the decrease in segmentation performance when encountering organs in isolation and removals becomes more pronounced across both modalities. For image-based segmentation, the performance drop is highest, ranging from \varRGBDropRange for RGB and \varHSIDropRange for \ac{hsi}, depending on the \ac{ood} scenario.

\begin{figure*}
    \centering
    \includegraphics[width=0.84\textwidth]{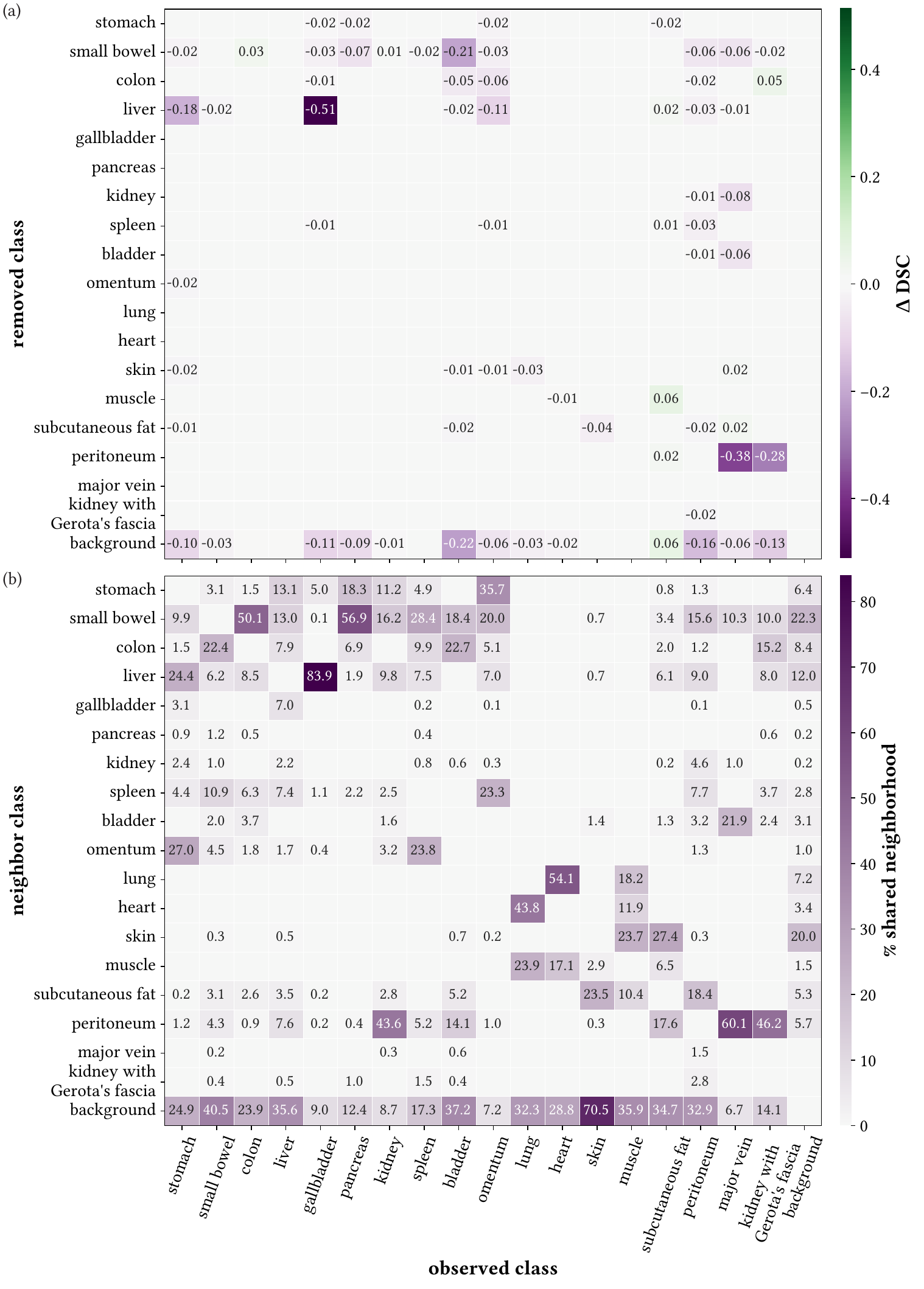}
    \caption{\textbf{Performance decline after organ removal is related to the local neighborhood.} (a) Performance difference of the image\#HSI model in response to class removals. The $i,j$-th entry displays the change in the average \acf*{dsc} of the $j$-th class when the $i$-th class is replaced with zeros. For clarity, values of $\left| \Delta \acs*{dsc} < 0.01 \right|$ are omitted. (b) Organ neighborhood matrix for the \textit{original} test dataset. The $(i,j)$-th entry indicates the average proportion of boundary pixels that the observed organ class $j$ shares with the organ class $i$. Values below \SI{0.1}{\percent} are not shown for clarity. For both subfigures, the aggregation was performed hierarchically by first averaging the proportions/scores across all images of one subject, and subsequently averaging across subjects. The performance matrix is adapted from \autocite{sellner_context_2023}.}
    \label{fig:neighbor_removal}
\end{figure*}

\autoref{fig:neighbor_removal} offers a detailed analysis of which organ classes are most impacted in the organ removal scenario for the image\#HSI model. The performance drop is highest for the gallbladder when the liver is removed, and second highest for the major vein when the peritoneum is removed. In both cases, the removed organ and the organ under investigation are frequent neighbors, with the liver comprising \varGallbladderLiverNeigbor of the gallbladder’s neighborhood, and the peritoneum accounting for \varMajorVeinPeritoneum of the major vein’s neighborhood, on average.

\subsection{Performance of Our Organ Transplantation Augmentation}
\label{sec:augmentation_results}

\autoref{fig:task_performance_dsc} and \autoref{fig:task_performance_nsd} demonstrate that our Organ Transplantation augmentation (gold) effectively addresses performance drops of image-based surgical scene segmentation models under geometric domain shifts for both the RGB and \ac{hsi} modalities. The performance improvement over the baseline ranges from \varRGBImprovementRangeDSC (\ac{dsc}) and \varRGBImprovementRangeNSD (\ac{nsd}) for RGB, and from \varHSIImprovementRangeDSC (\ac{dsc}) and \varHSIImprovementRangeNSD (\ac{nsd}) for \ac{hsi}. Even on in-distribution data, slight performance enhancements can be observed. The largest performance boost on \ac{ood} data is obtained in the isolation scenario.

\newcommand{\taskPerformanceDescription}[2]{\textbf{The proposed Organ Transplantation augmentation compensates for geometric domain shifts.} The dot and box plots show the segmentation performance of the baseline image model and our Organ Transplantation model for the modalities \acf*{hsi} (top) and RGB (bottom) across six geometric out-of-distribution datasets and two in-distribution datasets (highlighted in \textit{italic}). Each boxplot displays the \acf*{iqr} of the \acf*{#1} with the median (solid line) and mean (dotted line) and whiskers extending up to 1.5 times the \acs*{iqr}. Each point represents the average \acs*{#1} of one organ class. The aggregation was performed hierarchically by first averaging the \acs*{#1} values across all images of one subject, and subsequently averaging across subjects. Results for the \acf*{#2} are presented in \autoref{fig:task_performance_#2}. This figure is adapted from \autocite{sellner_context_2023}.}
\begin{figure*}[htb]
    \centering
    \includegraphics[width=\textwidth]{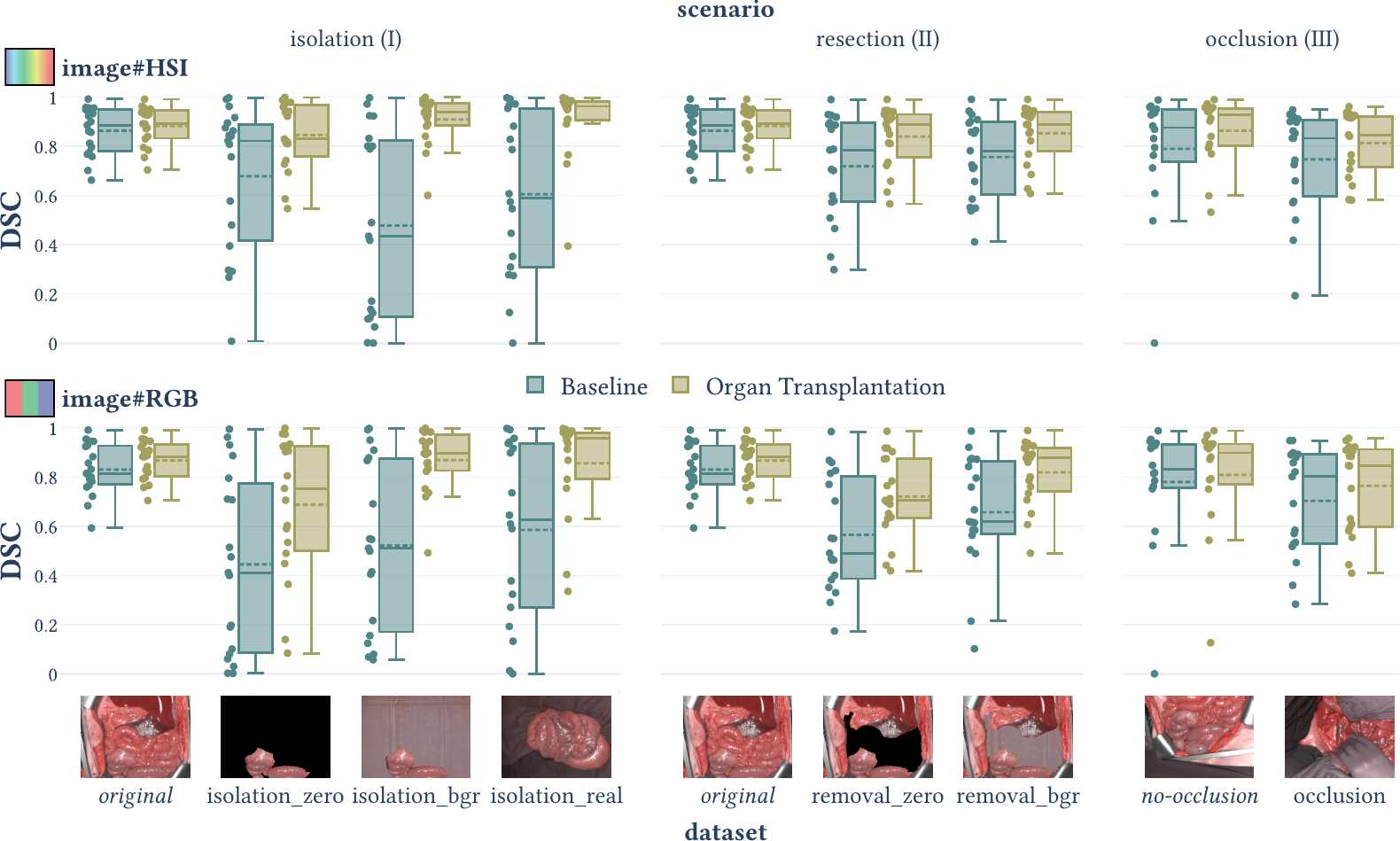}
    \caption{\taskPerformanceDescription{dsc}{nsd}}
    \label{fig:task_performance_dsc}
\end{figure*}

\begin{figure*}
    \centering
    \includegraphics[width=0.9\textwidth]{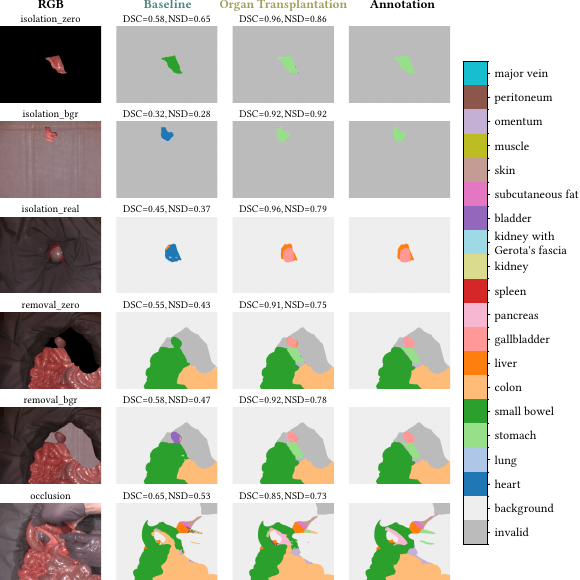}
    \caption{\textbf{Our augmentation drastically reduces segmentation errors even on the most severe failure cases.} Exemplary segmentation predictions are shown, comparing the baseline image\#HSI model to the corresponding Organ Transplantation augmentation model (columns) on six geometric out-of-distribution datasets (rows). For each prediction, performances measured by \acf*{dsc} and \acf*{nsd} are shown. Images were selected based on the maximum difference in \acs*{dsc} values between the baseline and Organ Transplantation augmentation models for each dataset.}
    \label{fig:example_predictions}
\end{figure*}

\autoref{fig:example_predictions} presents example predictions from the image\#HSI model with and without the Organ Transplantation augmentation for each \ac{ood} dataset. These images were selected based on the maximum difference in \ac{dsc} performance between the baseline model and the Organ Transplantation augmentation model, highlighting cases where the benefit from the augmentation is most significant. In all six examples, the performance drop in the baseline model is primarily due to mispredicting entire organ classes, such as mispredicting the gallbladder and stomach after liver removal, or failing to recognize the stomach and large parts of the omentum obstructed by a gloved hand in the occlusion scenario.

\subsection{Comparison to Other Topology-Altering Augmentations}
\label{sec:ranking}

\autoref{fig:ranking_dsc} displays the \ac{dsc}-based ranking of our Organ Transplantation augmentation compared to six other topology-altering augmentations on our geometric \ac{ood} test datasets. The \ac{nsd}-based ranking is presented in \autoref{fig:ranking_nsd}. Our Organ Transplantation augmentation consistently ranks first, whereas the baseline augmentations generally rank last in most \ac{ood} scenarios. Although the other ranks vary across different geometric \ac{ood} datasets, in the overall ranking, the image-mixing augmentations CutMix and Jigsaw outperform the noise augmentations Random Erasing and Hide-and-Seek.

\newcommand{\rankingDescription}[2]{\textbf{Our Organ Transplantation approach outperforms augmentation methods adapted from the general computer vision community.} The uncertainty-aware ranking of seven augmentation methods applied to the image\#HSI model on six geometric out-of-distribution datasets is shown. The area of each blob represents the relative frequency with which the corresponding algorithm achieved each rank across 1000 bootstrap samples, based on the \acf*{#1}. Each bootstrap sample consists of 19 hierarchically aggregated class-level \acs*{#1} metric values reflecting the variability across subjects. The lines encompass the \SI{95}{\percent} confidence interval of the bootstrap results, while the cross and the diamond markers denote the median and mean ranks, respectively. Results for the \acf*{#2} are shown in \autoref{fig:ranking_#2}. This figure is adapted from \autocite{sellner_context_2023}.}
\begin{figure*}[htb]
    \centering
    \includegraphics[width=\textwidth]{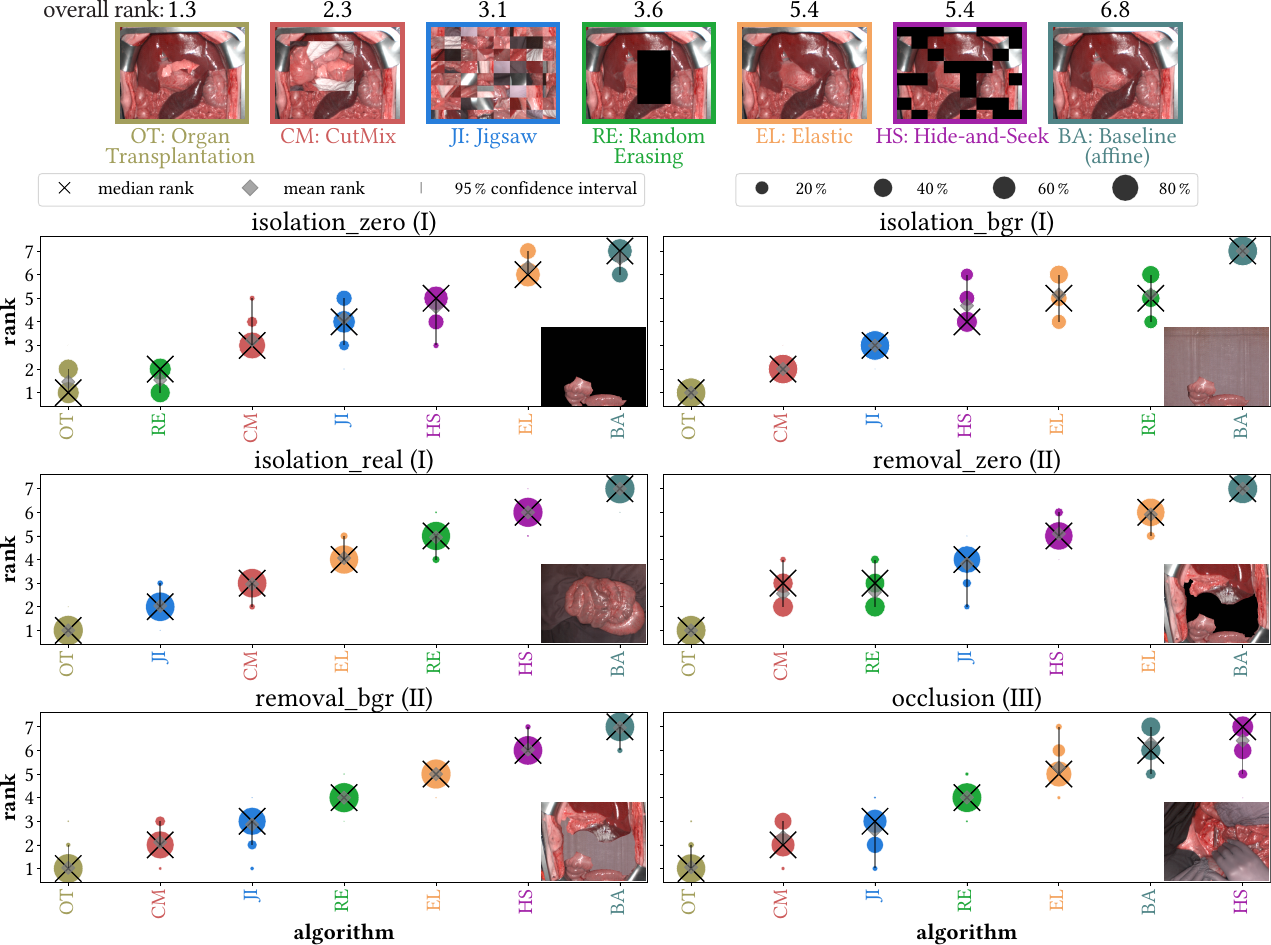}
    \caption{\rankingDescription{dsc}{nsd}}
    \label{fig:ranking_dsc}
\end{figure*}

\section{Discussion}
\label{sec:discussion}

In this study, we are the first to demonstrate that \ac{soa} surgical scene segmentation networks fail under geometric domain shifts. Through comprehensive validation on six geometric \ac{ood} datasets, comprising \varTotalImages RGB and \ac{hsi} cubes from \varTotalPigs pigs, each annotated with \varTotalClasses classes, we found that performance degradation was generally higher for RGB compared to \ac{hsi}. Additionally, the decline was more pronounced with increased input spatial granularity, such as images and patches, compared to smaller granularity like pixels and superpixels. To enhance the generalization of \ac{soa} models towards \ac{ood} geometries, we adapted previously unexplored topology-altering data augmentation methods to surgical scene segmentation. Our proposed Organ Transplantation augmentation outperformed all other topology-altering augmentations and achieved performance on par with in-distribution performance.

The following sections offer a detailed discussion of our experimental results (\autoref{sec:interpretation_results}), limitations and future work (\autoref{sec:limitations}), and our conclusions (\autoref{sec:conclusion}).

\subsection{Interpretation of Results}
\label{sec:interpretation_results}

\textit{Performance degradation across spatial granularities:}
We observed that performance degradation was more pronounced, the higher the input spatial granularity. However, some performance scores reported in \autoref{fig:spatial_granularities_dsc} should be interpreted with care, as there are artifacts due to limitations in our experiments: 
In the manipulated isolation scenarios, superpixel-based segmentation seemed to have improved over in-distribution data. This is due to our manipulation strategy. As demonstrated in \autocite{seidlitz_robust_2022}, superpixel boundaries on real data do not perfectly align with annotation boundaries, especially between different organ classes. However, on the manipulated data, we used the reference boundary annotation of the target organ to replace non-target pixels with zeros or background spectra. This resulted in superpixel boundaries that closely follow the annotations, leading to higher segmentation scores. Pixel-based segmentation also showed performance boosts on simulated and real isolation data. As demonstrated in \autocite{seidlitz_robust_2022}, pixel-based segmentation has incomplete and scattered boundaries between organs, whereas it accurately discriminates between tissue and background. In the simulated isolation scenarios, pixels outside the target organ annotation were replaced with zeros and background pixels, and in the \dataset{isolation\_real} dataset, the entire scene except for the target organ was covered by abdominal linen. This prevented the mispredictions of the target class outside the annotated area that typically occur in multi-organ images.

\textit{Which is the optimal spatial granularity?}
The optimal input spatial granularity was previously examined in \autocite{seidlitz_robust_2022} using in-distribution data from the \dataset{original} dataset, revealing that performance improves with larger spatial granularity. However, we demonstrated that \ac{soa} models with larger input spatial granularity were less robust to geometric domain shifts. Despite this, we believe image-based segmentation models are the best choice due to their training efficiency (e.g., augmentations can be computed on the GPU) \autocite{sellner_benchmarking_2023}, their superior segmentation results that could be restored through our Organ Transplantation augmentation, and their ability to apply batch-based augmentations that benefit from larger spatial context (cf. \autoref{sec:limitations}).

\textit{\ac{hsi} vs RGB:}
Previous work \autocite{seidlitz_robust_2022} showed that \ac{hsi}- outperforms RGB-based segmentation with in-distribution data. We found that \ac{hsi} also showed a smaller overall performance drop on geometric \ac{ood} data, underscoring the value of additional spectral information for handling geometric domain shifts.

\textit{Geometric \ac{ood} performance is class-dependent:}
Geometric \ac{ood} scenarios impacted different class labels unevenly, as demonstrated by the significant variability in class-level scores for \ac{ood} scenarios in \autoref{fig:task_performance_dsc}. For instance, in the removal scenario, performance dropped most notably for gallbladder and major vein when their frequent neighbor was removed, whereas liver, muscle, and background showed no performance decline with the removal of other classes.

\textit{Topology-altering augmentations:}
Our experiments were based on the hypothesis that topology-altering augmentations can introduce simulated geometric domain shifts during model training, thus enabling the models to generalize better on geometric \ac{ood} data. This hypothesis was confirmed, as in most scenarios, topology-altering augmentations outperformed the baseline augmentations. Overall, the image-mixing augmentations Organ Transplantation, CutMix, and Jigsaw outperformed the noise augmentations Random Erasing and Hide-and-Seek. This could be because image-mixing augmentations create unusual neighborhood relations by copying parts of one surgical scene onto another, whereas noise augmentations only obscure parts of the scene without altering neighborhood relations. Additionally, noise augmentations that black out parts of the scene may train models to handle this specific kind of obscuration, but not to deal with image parts being obscured by a known class. This might explain why Random Erasing ranked best on the \dataset{isolation\_zero} and \dataset{removal\_zero} datasets, where parts of the image are blacked out. We further observed that our Organ Transplantation augmentation performed best, and that CutMix and Random Erasing, which alter rectangular patches, outperformed their grid-based counterparts. This may be due to the level of unnatural boundaries introduced: Our Organ Transplantation augmentation preserves organ boundaries in unusual contexts, while CutMix and Random Erasing create unnatural boundaries at the edges of rectangles, and Jigsaw and Hide-and-Seek generate even more unnatural boundaries due to their grid-based modifications.

\textit{Value of manipulated data:}
We developed and optimized our augmentation models solely on the validation splits of the in-distribution dataset \dataset{original} and the manipulated \ac{ood} datasets \dataset{isolation\_zero} and \dataset{isolation\_bgr}, keeping all real-world \ac{ood} datasets as untouched test sets. Despite this, our proposed augmentation was effective on all datasets, demonstrating that image manipulations are a powerful tool for assessing geometric \ac{ood} performance. This is particularly valuable in scenarios with limited real-world data, such as organ resection, which would require an impractical number of animals.

\subsection{Limitations and Future Work}
\label{sec:limitations}

A general limitation of image-mixing augmentations like our Organ Transplantation method is that they require a minimum batch size of two images to transplant an organ from one scene to another. For image-based segmentation, this is not an issue, as batch sizes are typically larger (e.g., five images in our case). However, for other spatial granularities, implementing image-mixing augmentations would result in significant computational overhead and increased memory requirements: Since we believe that the preservation of organ boundaries is a key strength of our Organ Transplantation augmentation, it should be applied before extracting pixels, superpixels, or patches. As the latter must be done on the CPU due to memory and efficiency constraints, our augmentation would have to be run on the CPU as well. However, it has been shown that for efficient training, \ac{hsi} models should perform augmentations on the GPU instead \autocite{sellner_benchmarking_2023}. Additionally, the need to load at least two images simultaneously would substantially increase memory and load demands for lower spatial granularities, making batch-level augmentations impractical. For this reason, we only validated the topology-altering augmentations on image-based segmentation models. Given that image-based segmentation generally outperformed smaller spatial granularities and our method achieved geometric \ac{ood} performance comparable to in-distribution performance, using smaller spatial granularities would not provide additional benefits.

In this paper, we studied geometric \ac{ood} scenarios common in real-world open surgeries. Similar scenarios, such as instrument occlusions and organ removals, occur in minimally invasive surgeries. However, there are significant differences, including more close-up views of organs, fewer neighboring organs visible in an image, tissue deformations, and substantial changes in imaging perspectives. With the recent availability of medical device-graded \ac{hsi} systems for minimally invasive surgery and the steady increase in the number of such procedures over the past decades \autocite{john2020rise}, investigating and compensating for geometric domain shifts in this context is a promising future direction.

\subsection{Conclusion}
\label{sec:conclusion}

To the best of our knowledge, we are the first to address surgical scene segmentation under geometric domain shifts. We demonstrated significant performance drops in \ac{soa} segmentation models when faced with geometric \ac{ood} scenarios, but also showed that in-distribution performance could be restored using our Organ Transplantation augmentation. Our augmentation method stands out for being computationally efficient, effective, and model-independent, making it suitable for image-based surgical scene segmentation of both \ac{hsi} and RGB data across various architectures. Our code repository and pretrained models are publicly available in GitHub\footnote{\href{https://github.com/IMSY-DKFZ/htc}{https://github.com/IMSY-DKFZ/htc}} \autocite{sellner_htc_2023}, and we have also integrated our Organ Transplantation augmentation into the Kornia library\footnote{\href{https://kornia.readthedocs.io/en/stable/augmentation.module.html\#kornia.augmentation.RandomTransplantation}{Kornia \texttt{RandomTransplantation} augmentation}} \autocite{kornia_riba2020} for easy access by the broader computer vision community.

\printacronyms

\section*{Declaration of Competing Interest}
The authors declare that they have no known competing financial interests or personal relationships that could have appeared to influence the work reported in this paper. The Authors declare that there is no conflict of interest.

\section*{Declaration of generative AI and AI-assisted technologies in the writing process}
During the preparation of this work, the authors used ChatGPT in order to improve language. After using this tool/service, the authors reviewed and edited the content as needed and take full responsibility for the content of the published article.

\section*{Acknowledgments}
This project has received funding from the European Research Council (ERC) under the European Union’s Horizon 2020 research and innovation programme (project NEURAL SPICING grant agreement No. 101002198) and the National Center for Tumor Diseases (NCT) Heidelberg's Surgical Oncology Program. It was further supported by the German Cancer Research Center (DKFZ) and the Helmholtz Association under the joint research school HIDSS4Health (Helmholtz Information and Data Science School for Health). Lena Maier-Hein and Beat P. M\"uller-Stich worked with the medical device manufacturer KARL STORZ SE \& Co. KG in the projects \enquote{InnOPlan} and \enquote{OP 4.1}, funded by the German Federal Ministry of Economic Affairs and Energy (grant agreement No. BMWI 01MD15002E and BMWI 01MT17001E) and \enquote{Surgomics}, funded by the German Federal Ministry of Health (grant agreement No. BMG 2520DAT82D).

\bibliographystyle{model2-names.bst}
\biboptions{authoryear}
\bibliography{references,references_extended}

\appendix
\setcounter{figure}{0}

\section{Additional results}

\begin{figure*}
    \centering
    \includegraphics[width=\textwidth]{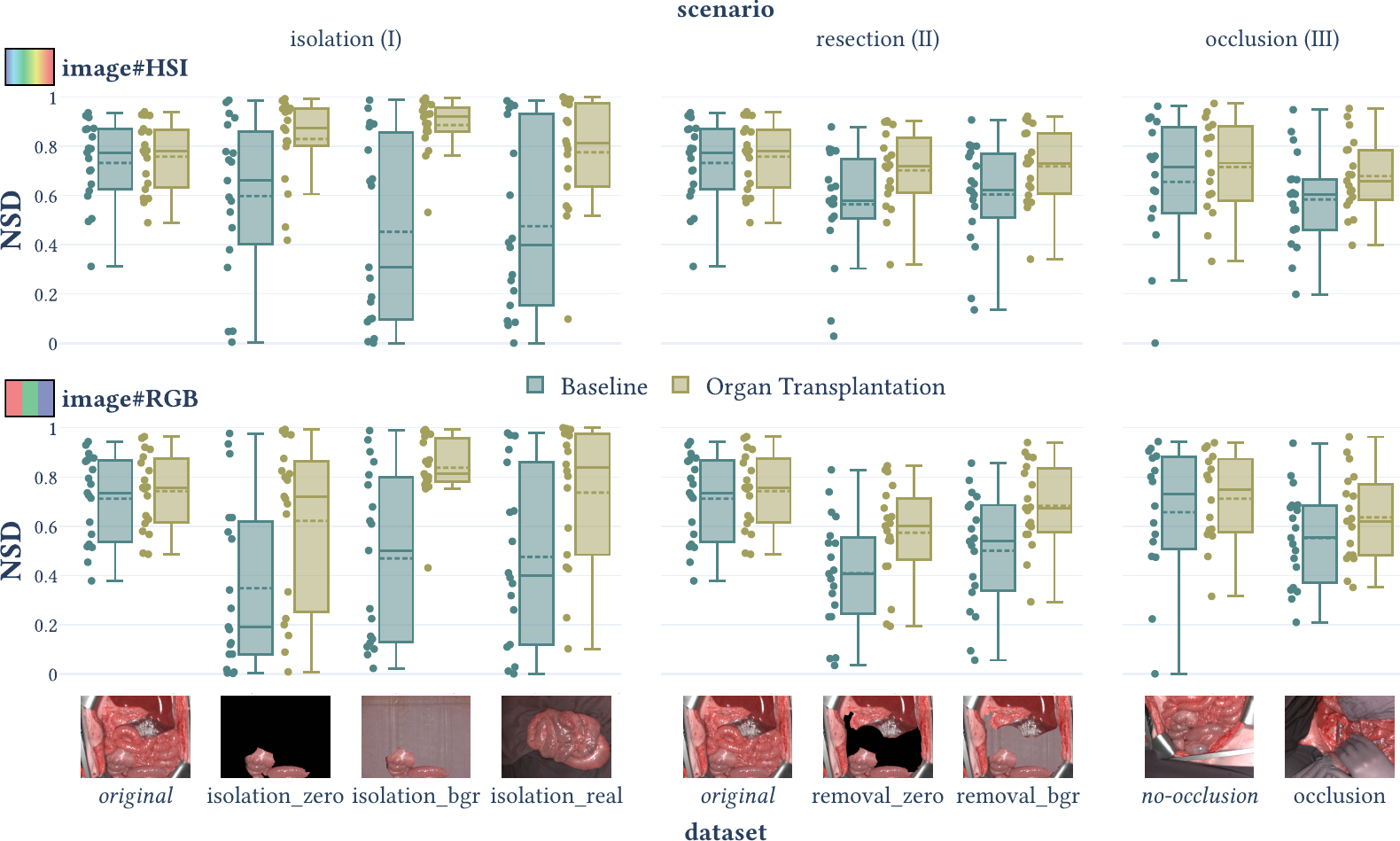}
    \caption{\taskPerformanceDescription{nsd}{dsc}}
    \label{fig:task_performance_nsd}
\end{figure*}

\begin{figure*}
    \centering
    \includegraphics[width=\textwidth]{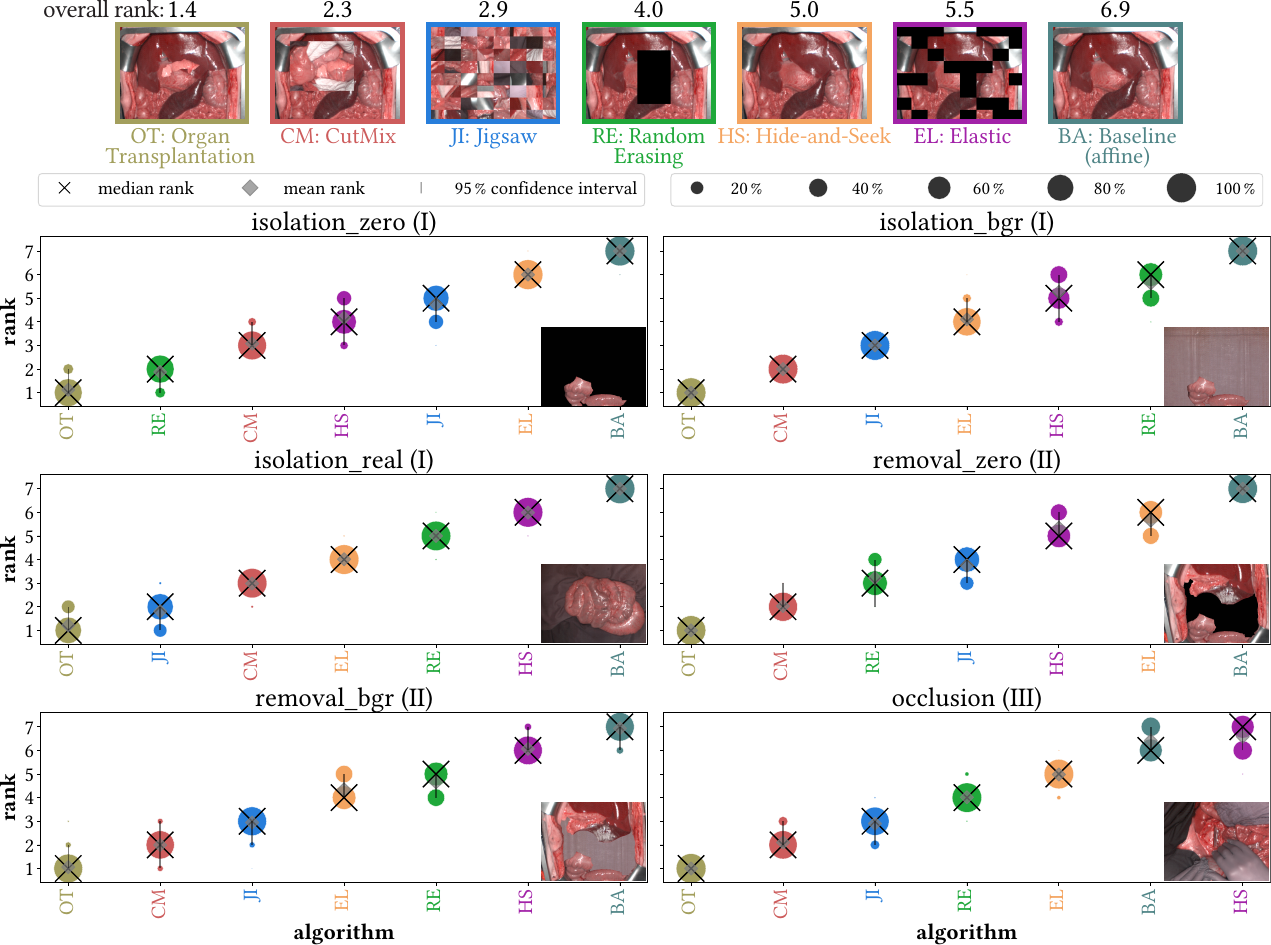}
    \caption{\rankingDescription{nsd}{dsc}}
    \label{fig:ranking_nsd}
\end{figure*}

\end{document}